%% file: iclr2021_conference.tex
\documentclass{article} 
\usepackage{iclr2021_conference,times}

\iclrfinalcopy
\usepackage{booktabs}
\usepackage{amsmath,amsthm,amsfonts,amssymb,amscd}
\usepackage{lastpage}
\usepackage{enumerate}
\usepackage{fancyhdr}
\usepackage{mathrsfs}
\usepackage{xcolor}
\usepackage{graphicx}
\usepackage{listings}
\usepackage{hyperref}
\usepackage{algorithm}
\usepackage{algorithmic}
\usepackage{braket}
\usepackage{caption}
\usepackage{subcaption}
\usepackage[algo2e,ruled,vlined]{algorithm2e}

\newcommand{\norm}[1]{\left\lVert#1\right\rVert}

\definecolor{airforceblue}{rgb}{0.36, 0.54, 0.66}

\hypersetup{%
  colorlinks=true,
  linkcolor=blue,
  citecolor=airforceblue,
  linkbordercolor={0 0 1}
}

\lstdefinestyle{Python}{
    language        = Python,
    frame           = lines, 
    basicstyle      = \footnotesize,
    keywordstyle    = \color{blue},
    stringstyle     = \color{green},
    commentstyle    = \color{red}\ttfamily
}

\input{math_commands.tex}

\usepackage{hyperref}
\usepackage{url}

\title{An Energy-Based View of Graph Neural Networks}

\author{%
  John Y.~Shin \\
  Dept. of Computer Science and Engineering\\
  New York University\\

  Brooklyn, NY 11201\\
  \texttt{jys308@nyu.edu} \\
  \And
   Prathamesh Dharangutte \\
    Dept. of Computer Science and Engineering\\
  New York University\\

  Brooklyn, NY 11201\\
  \texttt{ptd244@nyu.edu} \\
}

\begin{document}

\maketitle

\begin{abstract}
Graph neural networks are a popular variant of neural networks that work with graph-structured data. In this work, we consider combining graph neural networks with the energy-based view of \cite{energy} with the aim of obtaining a more robust classifier. We successfully implement this framework by proposing a novel method to ensure generation over features as well as the adjacency matrix and evaluate our method against the standard \emph{graph convolutional network} (GCN) architecture (\cite{kipf}). Our approach obtains comparable discriminative performance while improving robustness, opening promising new directions for future research for energy-based graph neural networks.
\end{abstract}

\section{Introduction}

Graph neural networks (GNNs) are a generalization of neural networks that operate on graph-structured data, typically in the form of an adjacency matrix or graph laplacian, and feature vectors defined for the nodes. They have found success in tasks such as link prediction, node classification, and graph classification (e.g., \cite{izadi2020optimization}, \cite{wang2019knowledge}, \cite{zhang2019hierarchical}). Alongside the empirical success, recent theoretical work has elucidated properties on their depth (\cite{expressive}), architectural alignment with algorithms (\cite{alignment}), and their discriminative power (\cite{powerful}). Recently, the work of \cite{energy} proposed viewing traditional classifiers as energy-based models, adjusting the softmax transfer function to the Boltzmann distribution, and adding an inner stochastic Langevin gradient descent (\cite{welling2011bayesian}) loop over new samples. We extend this framework for GNNs, and introduce a novel method to ensure generation over the adjacency matrix itself along with node features. We evaluate our approach for the node classification task on the Cora, Pubmed, and Citeseer datasets and explore its generative capabilities.

\section{Background and Prior Work}

One of the most popular GNN architectures is the GCN architecture of \cite{kipf}, which utilizes a normalized adjacency matrix with self-connections. Consider the adjacency matrix of a graph, $\mA \in \mathbb{R}^{n \times n}$. We define $\hat{\mA} = \mA + \mI_n$, where $\mI_n$ is the $n \times n$ identity matrix. Let $\hat{\mD}_{ii} \equiv \sum_{j}\hat{\mA}_{ij}$. Layer-wise propagation is given by:

\begin{equation}
\label{eq:gcn}
\mH^{l+1} = \sigma(\hat{\mD}^{-1/2} \hat{\mA} \hat{\mD}^{-1/2}\mH^l \mW^l)
\end{equation}

Equation \ref{eq:gcn} follows from \cite{kipf}, and multiplication by $\hat{\mD}^{-1/2}$ ensures symmetric normalization of the graph adjacency matrix $\hat{\mA}$. $\mH^0 = \mX \in \mathbb{R}^{n \times f}$ is the data matrix with row-wise examples, where $f$ is the number of features. $\sigma(\cdot)$ is a non-linear activation function, and $\mW^l$ is a linear transformation which typically reduces the dimensionality of the data. The output of the final layer is $k$-dimensional, where $k$ is the number of classes. This is passed through a softmax function, which is then optimized through the cross-entropy loss.

In the work of \cite{energy}, it was noted that the softmax function can be changed in the following way: with typical machine learning classification, we are interested in a function $f_{\theta}(\vx)$, which maps the input data $\vx \in \mathbb{R}^{d} \rightarrow \mathbb{R}^k$ to $k$ real valued numbers, where $k$ is the number of classes, and $\theta$ are trainable parameters. When this is passed through a softmax function, this can be viewed as a  conditional distribution of $y$ conditioned on $\vx$. In viewing this as a conditional distribution, we can also see that the joint distribution is given by normalization by the partition function.
\begin{equation}
p_{\theta}(y | \vx) = \frac{\exp(f_{\theta}(\vx)[y])}{\sum_y \exp(f_{\theta}(\vx)[y])} \rightarrow p_{\theta}(\vx,y) = \frac{\exp(f_{\theta}(\vx)[y])}{Z(\theta)}
\end{equation}
By marginalization of $y$, we have the probability density function of $\vx$:
\begin{equation}
p_{\theta}(\vx) = \frac{\sum_y \exp(f_{\theta}(\vx)[y])}{Z(\theta)}
\end{equation}
By viewing this as a Boltzmann distribution, the energy function is given as:
\begin{equation}
E_{\theta}(\vx) = -\log \left(\sum_y \exp(f_{\theta}(\vx)[y])\right)
\end{equation}

An energy-based view of convolutional neural networks was given in the work of \cite{xie2016theory}, and applied to various tasks in subsequent works (\cite{xie2017synthesizing, xie2019learning}). Combining energy-based methods and graph neural networks has been explored in the works of \cite{cao2020energy} and \cite{ma2019genn}. In \cite{ma2019genn}, the energy is given as the sum over the node embeddings produced by a graph neural network normalized by the number of nodes, which is input into a multilayer perceptron and passed through a ReLU. In \cite{cao2020energy}, an analogy to the coulomb potential is used in designing a feature pooling mechanism. Our work focuses on extending the framework of \cite{energy} for GNNs and the exact approach is detailed in the following section.



\section{Methodology}



In essence, the algorithm is a stochastic Langevin gradient descent (SLGD) loop nested inside of an stochastic gradient descent (SGD) loop. The inner SLGD loop samples the generative features over some simple distribution, and minimizes the energy of the new features using SGLD. The outer SGD loop computes the cross-entropy of the classifier, and combines the loss for the classifier as well as the generative loss, and computes the gradient. Finally, within the SGD loop, we generate new links in the adjacency matrix for nodes that are close in \emph{energy}. 

\begin{algorithm}[h]
\SetAlgoLined
\KwResult{Trained network $f_{\theta}$. Generative features, $\hat{\vx}_t$ in $\sB$. Generative graph adjacency, $\tilde{\mA}$.  Feature matrix with generative samples, $\hat{\mX}$. }\;
 \While{not converged}{
    $L_{\text{clf}}(\theta) = \text{xent}(f_{\theta}(\mA, \mX), y)$\; \\
    Batch sample $\vx^i$ and $y^i$ from dataset $\mX$, with number of samples $\zeta$. Let this set be $\sS$. \\
    \For{all $i \in \sS$}{
    Sample $\hat{\vx}_0^i \sim \sB$ with probability $1-\rho$, else $\hat{\vx}_0^i \sim \mathcal{U}(-1,1)$.\\
    Replace $\vx^i$ with $\hat{\vx}^i_0$ in $\mX$, creating $\hat{\mX}_0$. Also replace $\vx^i$ with $\hat{\vx}^i_0$ in $\sS$.\\
    \For{$t \in [0, \hdots, \eta]$}{
        Let $Z_t = \text{LogSumExp}_{y'}f_{\theta}(\mA, \hat{\mX}_t)[y']$ \hspace{20pt}  (energy of the graph with the new features)\\
        $\hat{\vx}_{t+1}^i = \hat{\vx}_{t}^i + \alpha \frac{\partial \text{LogSumExp}_{y'}(f_{\theta}(\mA, \hat{\vx}_{t}^i)[y'])/Z_t}{\partial \hat{\vx}_{t}^i} + \sigma \mathcal{N}(0, I)$
        }
    }
    
    Let $Z_{\text{gen}} = \text{LogSumExp}_{y'}f_{\theta}(\mA, \hat{\mX})[y']$ \hspace{25pt} (energy of the graph with the new features)\;\\
    Let $Z_{\text{clf}} = \text{LogSumExp}_{y'}f_{\theta}(\mA, \mX)[y']$ \hspace{20pt} (energy of the graph with the original features)\;\\
    $L_{\text{gen}}(\theta) = | \sum_{i \in \sS}\text{LogSumExp}_{y'}(f_{\theta}(\mA, \vx^i)[y'])/Z_{\text{clf}} - \text{LogSumExp}_{y'}(f_{\theta}(\mA, \hat{\vx}^i_t)[y'])/Z_{\text{gen}}|$\\
    $L(\theta) = L_{\text{clf}}(\theta) + L_{\text{gen}}(\theta)$\;\\
    
    Compute gradients $\frac{\partial L (\theta)}{\partial \theta}$ and propagate.\;\\
    Add $\sS$ to $\sB$.
    
    \For{All the new features, $\hat{\vx}_t^i$ in $\sS$}{
        \If {$|\text{LogSumExp}_{y'}(f_{\theta}(\mA, \hat{\vx}_{t}^i)[y']) - \text{LogSumExp}_{y'}(f_{\theta}(\mA, \hat{\vx}_{t}^j)[y'])| \leq \tau $}
        {Add edge between nodes $i$ and $j$ in adjacency matrix, $\tilde{\mA}$.}
    }
    Set $\mA = \tilde{\mA}$\;
 }
 \caption{GCN-JEM training: Initialize $f_{\theta}$, SLGD step-size $\alpha$. SLGD noise $\sigma$, replay buffer $\sB$, SGLD steps $\eta$, reinitialization frequency $\rho$, energy threshold $\tau$. The original feature matrix $\mX$ with samples $\vx$ and labels $y$. Batch sample size $\zeta$. Batch sample set $\sS$. }
 \label{algorithm}
\end{algorithm}

The key factor for convergence when using stochastic Langevin gradient descent training with graph structured data is the renormalization by the total graph energy. The intuitive justification is that the gradient correction for the features of a node takes into account the total energy of the graph, since the messages are passed over the entire graph in the forward pass. The generative loss, $L_{gen}$, ensures that the energy of the original classifier over the selected indices is close to the energy of the generative features. This encourages their probabilities to be close to each other.

In addition to the generation over features, we have incorporated generation over the adjacency matrix itself. The motivation for this stems from the fact that for traditional machine learning models, samples are thought to be independent but that need not be the case for graph structured data as nodes are connected by edges. We use the energy of a sample as a quantitative measure to incorporate edges between the sampled data and existing graph. We compare the energy of two nodes, and add a link to the graph if the difference between the energy of two nodes is within the threshold $\tau$. For stability, we do this once every 50 epochs. In the complex networks literature (e.g., \cite{krioukov2010hyperbolic}), which has connections to the statistical mechanics literature, the connection probability of two nodes can be given by the Fermi-Dirac distribution. Under this assumption, two nodes that are close in energy have a high probability of being connected.


\section{Experimental results}

To evaluate our approach for training an energy-based GNN, three citation network datasets are considered -- Cora, Pubmed and Citeseer. In Table \ref{accuracy}, the reported accuracies are the average performance on the test set for 5 runs with random initialization, with a train-test split following that of \cite{kipf}. The deep graph library (DGL) framework (\cite{wang2019dgl}) is utilized for implementing the GCN upon which the energy based framework was implemented. The hyperparameters used in the experiments are given in Table \ref{params}. GCN-JEMO corresponds to the model with an additional term in the loss function which encourages the weight matrix $\mW^{l}$ for each layer $l$ to be orthogonal. We observe better performance with this additional constraint that promotes weight orthogonality:

\begin{equation}
\label{eq:ortho}
\norm{(\mW^{l})^T\mW^{l} - I}_F
\end{equation}

where  $\|.\|_F$ denotes the standard Frobenius norm $(\| \mA\|_F^2 = \sum_{ij} \mA_{ij}^2)$.

\begin{table}[!htbp]
    \centering
\begin{tabular}{llll}
    \toprule
    Model & CORA & Pubmed & Citeseer \\ 
    \midrule
    GCN & 81.5 & \textbf{79.0} & \textbf{70.3} \\
    GCN-JEM (ours) & 82.44&  77.04 & 67.28\\
    GCN-JEMO (ours) & \textbf{83.66}&  77.6 & 66.72\\ 
    \bottomrule
\end{tabular}
\end{table}

It should be noted that some papers for the Cora and PubMed datasets appear to use different training/test splits than the one used in the original GCN paper. To be consistent with the literature, we use the original train/test splits in the GCN paper (\cite{kipf}).

Our technique is in a sense \emph{architecture agnostic} and can be used with any graph neural network with a similar message-passing scheme. In our study, we use the \emph{vanilla} GCN (\cite{kipf}) as a baseline comparison. In theory, we could adapt our technique with any of the leading architectures.

\begin{table}[!htbp]
    \centering
\begin{tabular}{ll} 
    \toprule
    Parameter & Value \\
    \midrule
    epoch & 500  \\ 
	learning rate & 0.01  \\ 
    SGLD lr & 1\\ 
	SGLD steps & 20 \\ 
	 $\rho$ & 0.05 \\ 
	SGLD noise & 0.01 \\ 
	Sampling batch size & 32 \\ 
	\bottomrule
\end{tabular}
\caption{Hyperparameters used for experiments}\label{params}
\end{table}

\section{Discussion}

Our experimental results do not consistently improve the accuracy across the three datasets (an improvement for CORA is obtained), but it should be noted that the main focus of \cite{energy} was to improve the robustness of classifiers while having discriminative performance comparable to other top performing models. In \cite{energy}, the authors compare their method to other hybrid models, and in fact lose accuracy compared to a purely discriminative model. As such, the generative capabilities of the proposed model is explored.

\begin{figure}[h!]
    \centering
\begin{subfigure}{.5\textwidth}
  \centering
  \includegraphics[scale=0.5]{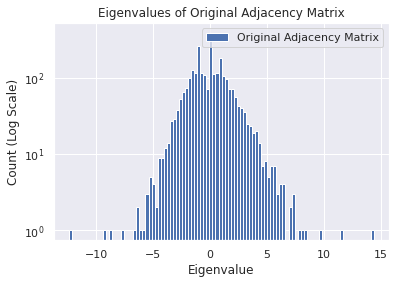}
  \caption{Original $\mA$ matrix}
  \label{fig:sub1}
\end{subfigure}%
\begin{subfigure}{.5\textwidth}
  \centering
  \includegraphics[scale=0.5]{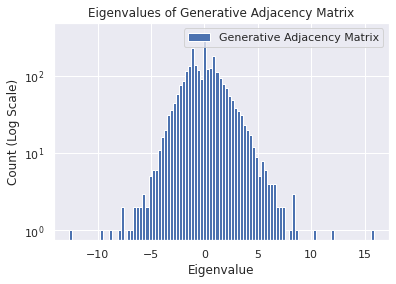}
  \caption{Generative $\tilde{\mA}$ matrix}
  \label{fig:sub2}
\end{subfigure}
    \caption{Spectrum of the original adjacency matrix for the Cora dataset compared to the generative adjacency matrix. The $x$-axis is the eigenvalue, and the $y$-axis is the count with log scaling. While the spectrum of both appear to be similar, the number of short cycles in the generative $\tilde{\mA}$ has increased.}
    \label{adjacency}
\end{figure}

\textbf{Spectrum of the adjacency matrix: } In Figure \ref{adjacency}, the spectrum of the original adjacency matrix and the spectrum of the generative adjacency matrix for the Cora dataset is shown. The $x$-axis is the eigenvalue, and the $y$-axis is the count of the eigenvalue with log scaling. The number of bins is set at $100$. It is seen that the spectrum of the generative adjacency matrix resembles that of the spectrum of the original adjacency matrix. By only adding links to nodes that are close in energy, it is thought that the generative $\Tilde{\mA}$ is only adding short cycles. The number of closed paths of length $n$ can be computed with the following formula:

\begin{equation}
\{\#\text{closed paths of length $n$}\} = \sum_i \lambda_i^n
\end{equation}

Where $\lambda_i, i \in \{1 , \cdots, n \}$ is the spectrum of the adjacency matrix. The cycles of length $n=3$ are computed at $9780$ for the original matrix, and $10674$ for the generative matrix.

\begin{figure}[h!]
    \centering
\begin{subfigure}{.5\textwidth}
  \centering
  \includegraphics[scale=0.5]{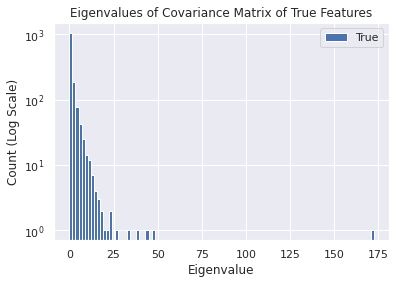}
  \caption{Spectrum of the Covariance of Original Features}
  \label{fig:sub3}
\end{subfigure}%
\begin{subfigure}{.5\textwidth}
  \centering
  \includegraphics[scale=0.5]{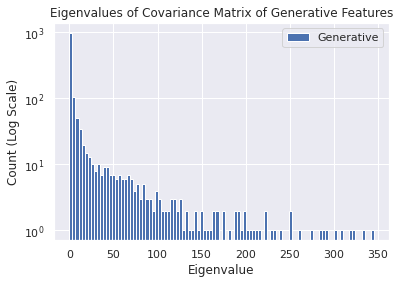}
  \caption{Spectrum of the Covariance of Generative Features}
  \label{fig:sub4}
\end{subfigure}
    \caption{Spectrum of the covariance matrix of the original and generative features for the Cora dataset. The $x$-axis is the eigenvalue, and the $y$-axis is the count with log scaling. The generative features have many more high eigenvalues.}
    \label{features}
\end{figure}

\textbf{Spectrum of the feature covariance matrix:} In Figure \ref{features}, the spectrum of the covariance of the original features as well as the covariance of the generative features for the Cora dataset are shown. The $x$-axis is the eigenvalue, and the $y$-axis is the count with log scaling. The number of bins is set at $100$. The generative features add high covariance eigenvalues, which may add robustness against adversarial examples. It is hypothesized that if the features have more variance, then the classifier is less likely to be overtuned to the dataset.

\begin{figure}[h!]
    \centering
\begin{subfigure}{.5\textwidth}
  \centering
  \includegraphics[scale=0.5]{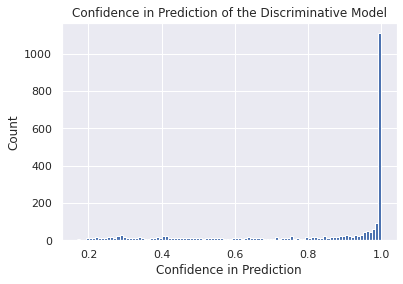}
  \caption{Confidence of Discriminative Model}
  \label{fig:sub5}
\end{subfigure}%
\begin{subfigure}{.5\textwidth}
  \centering
  \includegraphics[scale=0.5]{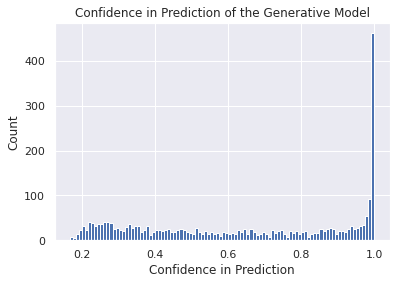}
  \caption{Confidence of Generative Model}
  \label{fig:sub6}
\end{subfigure}
    \caption{Histogram of the confidence in predictions in the training data for the Cora dataset. The $x$-axis is the confidence of the model on an example, where the confidence is the max of the softmax output. The $y$-axis is the count on a normal scale. Overall, the generative model is less confident.}
    \label{confidence}
\end{figure}

\textbf{Confidence in predictions: } In Figure \ref{confidence}, the confidence of the prediction over the training dataset is shown for both the discriminative and generative models, where confidence is the max of the softmax of the output. The $x$-axis is the confidence in the prediction, and the $y$-axis is the count at regular scaling. The number of bins is set to $100$. It is seen that the generative model overall is less confident in its predictions. 

One metric to measure the calibration of a classifier is the Expected Calibration Error (ECE). First, one computes the confidence in the predictions of the dataset. Then, one bins these into equally spaced buckets. The ECE measures the absolute difference between the average accuracy over that bucket and the average confidence, weighed by the cardinality of the bucket over the total number of samples. For a perfectly calibrated classifier, this value will be 0 for any choice of $M$, the number of buckets.

\begin{equation}
ECE = \sum_{m=1}^M \frac{|B_m|}{n}|\text{acc}(B_m) - \text{conf}(B_m)|
\end{equation}

We compute an ECE of $0.52$ for the generative model and $0.67$ for the discriminative model over the test set of Cora (lower is better).

\subsection{Future directions}
Combining the energy-based framework with graph neural networks presents promising new directions for further research, both from a  graph generation and theoretical point-of-view. One point of further exploration is exploring the limitations of depth as set out in \cite{expressive}. The paper posits that GNNs are limited in depth due to the action of repeated application of the graph operator. It is thought that this can be ameliorated by perturbing $\mA$ at every layer within the generative framework proposed in this work. Another direction for future work would be in utilizing our generative model for tasks in computational chemistry and bioinformatics tasks, where graph structure is abound, and where many tasks are inverse problems--given some chemical property that one requires, what is the structure that gives that property? Such tasks are thought to be apt for a generative graph model.



\bibliography{iclr2021_conference}
\bibliographystyle{iclr2021_conference}

\end{document}

%% file: math_commands.tex
\usepackage{amsmath,amsfonts,bm}

\def\eqref#1{equation~\ref{#1}}

\def\1{\bm{1}}

\def\vx{{\bm{x}}}

\def\mA{{\bm{A}}}

\def\mD{{\bm{D}}}

\def\mH{{\bm{H}}}
\def\mI{{\bm{I}}}

\def\mW{{\bm{W}}}
\def\mX{{\bm{X}}}

\DeclareMathAlphabet{\mathsfit}{\encodingdefault}{\sfdefault}{m}{sl}
\SetMathAlphabet{\mathsfit}{bold}{\encodingdefault}{\sfdefault}{bx}{n}

\def\sB{{\mathbb{B}}}

\def\sS{{\mathbb{S}}}

